\def\year{2019}\relax
\documentclass[letterpaper]{article} 
\usepackage{aaai19}  
\usepackage{times}  
\usepackage{helvet}  
\usepackage{courier}  
\usepackage{url}  
\usepackage{graphicx}  
\usepackage{amsmath}
\usepackage{amssymb}
\usepackage{multirow}
\usepackage{enumitem}
\usepackage{colortbl}

\begin{document}
%
\title{Long Short-Term Memory with Dynamic Skip Connections}
\author{Tao Gui, Qi Zhang, Lujun Zhao, Yaosong Lin, Minlong Peng, Jingjing Gong, Xuanjing Huang\\
Shanghai Key Laboratory of Intelligent Information Processing, Fudan University\\
School of Computer Science, Fudan University\\
Shanghai Insitute of Intelligent Electroics \& Systems\\
825 Zhangheng Road, Shanghai, China\\
\{tgui16, qz, ljzhao16, mlpeng16, yslin18, jjgong, xjhuang\}@fudan.edu.cn
}
\maketitle
\begin{abstract}
In recent years, long short-term memory (LSTM) has been successfully used to model sequential data of variable length. However, LSTM can still experience difficulty in capturing long-term dependencies. In this work, we tried to alleviate this problem by introducing a dynamic skip connection, which can learn to directly connect two dependent words. Since there is no dependency information in the training data, we propose a novel reinforcement learning-based method to model the dependency relationship and connect dependent words. The proposed model computes the recurrent transition functions based on the skip connections, which provides a dynamic skipping advantage over RNNs that always tackle entire sentences sequentially. Our experimental results on three natural language processing tasks demonstrate that the proposed method can achieve better performance than existing methods. In the number prediction experiment, the proposed model outperformed LSTM with respect to accuracy by nearly 20\%. 
\end{abstract}

\section{Introduction}
Recurrent neural networks (RNNs) have achieved significant success for many difficult natural language processing tasks, e.g., neural machine translation~\cite{sutskever2014sequence}, conversational/dialogue modeling~\cite{serban2016building}, document summarization~\cite{nallapati2016abstractive}, sequence tagging~\cite{santos2014learning}, and document classification~\cite{dai2015semi}. Because of the need to model long sentences, an important challenge encountered by all these models is the difficulty of capturing long-term dependencies. In addition, training RNNs using the ``Back-Propagation Through Time'' (BPTT) method is vulnerable to vanishing and exploding gradients.

\begin{figure}[t]
\centering
  \includegraphics[width=3in]{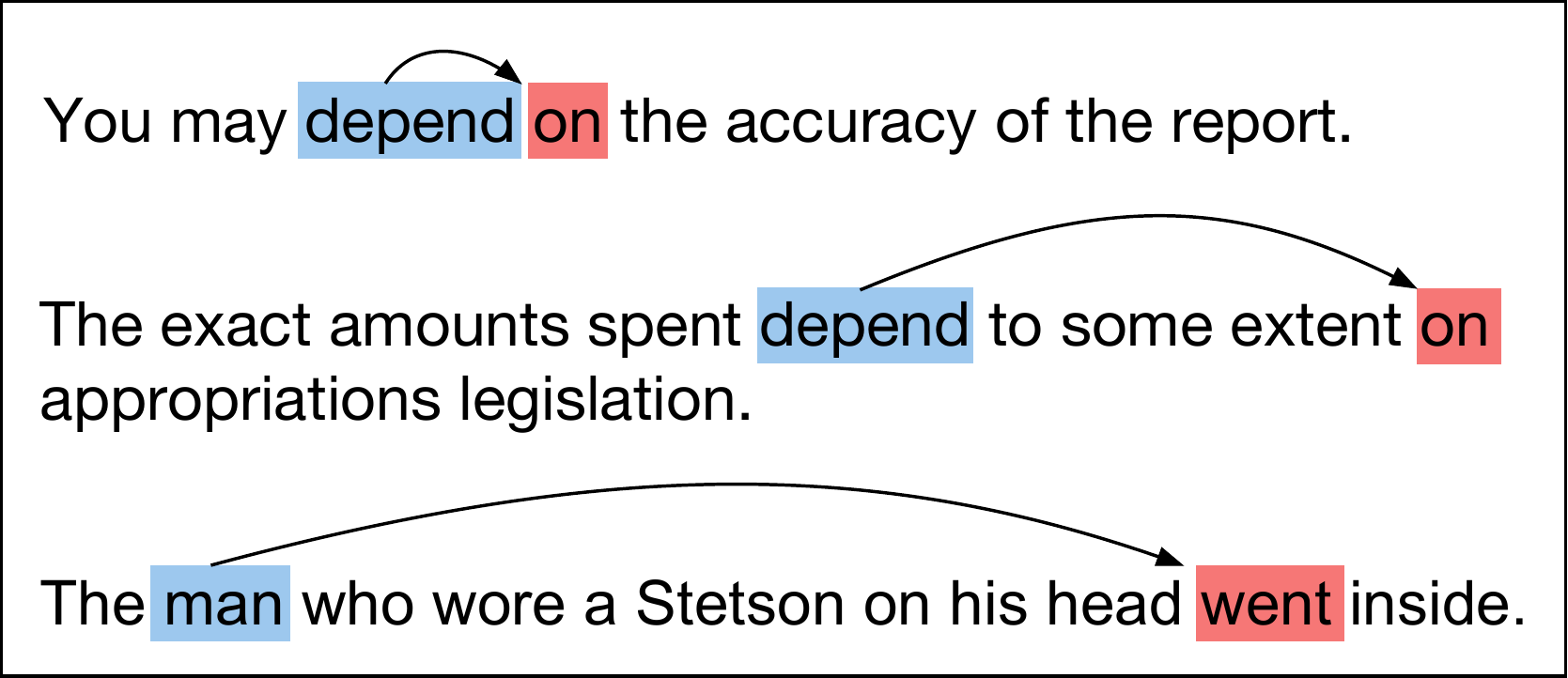}
  \caption{Examples of dependencies with variable length in the language. The same phrase ``\textit{depend on}'' in different sentences would have dependencies with different lengths. The use of clauses also makes the dependency length uncertain. Therefore, the models using a plain LSTM or an LSTM with fixed skip connections would be difficult to capture such information.} \label{fig:example}
\end{figure}

\begin{figure*}[t!]
\centering
\includegraphics[width=6.0in]{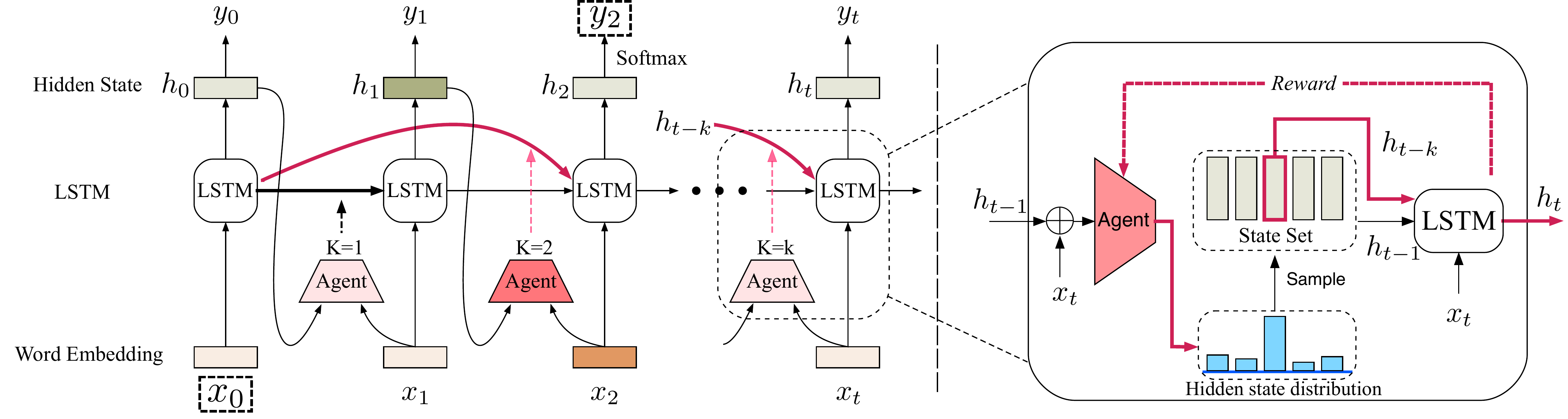}
\caption{Architecture of the proposed model. At time step $t$, the agent selects one of the past few states based on the current input $x_t$ and the previous hidden state $h_{t-1}$. The agent's selections will influence the log-likelihood of the ground truth, which will be a reward or penalty to optimize the agent. Take the phrase ``\textit{depend to some extent  on}'' as an example, the agent should learn to select the hidden state from ``\textit{depend}'' not ``\textit{extend}'' to predict ``\textit{on},'' because selecting ``\textit{depend}'' receives a larger reward.} \label{fig:model}
\end{figure*}

To tackle the above challenges, several variations of RNNs have been proposed using new RNN transition functional units and optimization techniques, such as gated recurrent unit (GRU)~\cite{chung2014empirical} and long short-term memory (LSTM)~\cite{hochreiter1997long}. Recently, many of the existing methods have focused on the connection architecture, including ``stacked RNNs''~\cite{el1996hierarchical} and ``skip RNNs''~\cite{chang2017dilated}. \citeauthor{zhang2016architectural} \shortcite{zhang2016architectural} introduced a general formulation for RNN architectures and proposed three architectural complexity measures: recurrent skip coefficients, recurrent depth, and feedforward depth. In addition, the skip coefficient is defined as a function of the “shortest path” from one time to another. In particular, they found empirical evidence that increasing the feedforward depth might not help with long-term dependency tasks, while increasing the recurrent skip coefficient could significantly improve the performance on long-term dependency tasks.

However, these works on recurrent skip coefficients adopted fixed skip lengths~\cite{zhang2016architectural,chang2017dilated}. Although quite powerful given their simplicity, the fixed skip length is constrained by its inability to take advantage of the dependencies with variable lengths in the language, as shown in Figure~\ref{fig:example}. From this figure, we can see that the same phrase ``\textit{depend on}'' in different sentences would have dependencies with different lengths. The use of clauses also makes the dependency length uncertain. In addition, the meaning of a sentence is often determined by words that are not very close. For example, consider the sentence ``\textit{The man who wore a Stetson on his head went inside.}'' This sentence is really about a man going inside, not about the Stetson. Hence, the models using a plain LSTM or an LSTM with a fixed skip would be difficult to capture such information and insufficient to fully capture the semantics of the natural language.

To overcome this limitation, in this paper, we consider the sequence modeling problem with dynamic skip connections. The proposed model allows ``LSTM cells'' to compute recurrent transition functions based on one optimal set of hidden and cell states from the past few states. However, in general, we do not have the labels to guide which two words should be connected. To overcome this problem, we propose the use of reinforcement learning to learn the dependent relationship through the exploring process. The main benefit of this approach is the better modeling of dependencies with variable length in the language. In addition, this approach also mitigates vanishing and exploding gradient problems with a shorter gradient backpropagation path. Through experiments, We find the empirical evidences (see Experiments and Appendix) that our model is better than that using attention mechanism to connect two words, as reported in~\cite{deng2018latent}. Experimental results also show that the proposed method can achieve competitive performance on a series of sequence modeling tasks.

The main contributions of this paper can be summarized as follows: 1) we study the sequence modeling problem incorporating dynamic skip connections, which can effectively tackle the long-term dependency problems; 2) we propose a novel reinforcement learning-based LSTM model to achieve the task, and the proposed model can learn to choose one optimal set of hidden and cell states from the past few states; and 3) several experimental results are given to demonstrate the effectiveness of the proposed method from different aspects.

\section{Approach}
In this work, we propose a novel LSTM network, which is a modification to the basic LSTM architectures. By using dynamic skip connections, the proposed model can choose an optimal set of hidden and cell states to compute recurrent transition functions. For the sake of brevity, we use $State$ to represent both the hidden state and cell state. Because of the non-differentiability of discrete selection, we adopted reinforcement learning to achieve the task.


\subsection{Model Overview}
Taking language modeling as an example, given an input sequence $x_{1:T}$ with length $T$, at each time step $t$, the model takes a word embedding $\mathbf{x}_t$ as input, and aims to output a distribution over the next word, which is denoted by $y_t$. However, in the example shown in Figure~\ref{fig:model}, in standard RNN settings, memorizing long-term dependency (depend ...$\mapsto$on) while maintaining short-term memory (to$\mapsto$some$\mapsto$extent) is difficult~\cite{chang2017dilated}. Hence, we developed a skipping technique that learns to choose the most relevant $State$ at time step $t-3$ to predict the word ``on'' to tackle the long-term dependency problem.

The architecture of the proposed model is shown in Figure~\ref{fig:model}. At time step $t$, the agent takes previous hidden state $h_{t-1}$ as input, and then computes the skip softmax that determines a distribution over the skip steps between 1 and $K$. In our setting, the maximum size of skip $K$ is chosen ahead of time. The agent thereby samples from this distribution to decide which $State$ is transferred to a standard LSTM cell for recurrent transition computation. Then, the standard LSTM cell will encode the newly selected $State$ and $x_t$ to the hidden state $h_t$. At the end of each time step, each hidden state $h_t$ is further used for predicting the next word based on the same method as standard RNNs. Especially, such a model can be fully applied to any sequence modeling problem. In the following, we will detail the architecture and the training method of the proposed model.


\subsection{Dynamic Skip with REINFORCE}
The proposed model consists of two components: (1) a policy gradient agent that repeatedly selects the optimal $State$ from the historical $State$ set, and (2) a standard LSTM cell~\cite{hochreiter1997long} using a newly selected $State$ to achieve a task. Our goal for training is to optimize the parameters of the policy gradient agent $\theta_a$, together with the parameters of standard LSTM and possibly other parameters including word embeddings denoted as $\theta_l$. 

The core of the proposed model is a policy gradient agent. Sequential words $x_{1:T}$ with length $T$ correspond to the sequential inputs of one episode. At each time step $t$, the agent interacts with the environment $s_t$ to decide an action $a_t$ (transferring a certain $State$ to a standard LSTM cell). Then, the standard LSTM cell uses the newly selected $State$ to achieve the task. The model's performance based on the current selections will be treated as a reward to update the parameters of the agent. 

Next, we will detail the four key points of the agent, including the environment representation $s_t$, the action $a_t$, the reward function, and the recurrent transition function.

\noindent \textbf{Environment representation.} Our intuition in formulating an environment representation is that the agent should select an action based on both the historical and current information. We therefore incorporate the previous hidden state $h_{t-1}$ and the current input $x_t$ to formulate the agent's environment representation as follows: $$s_t = h_{t-1} \oplus x_t,$$
where $\oplus$ refers to the concatenation operation. At each time step, the agent observes the environment $s_t$ to decide an action.

\noindent \textbf{Actions.} After observing the environment $s_t$, the agent should decide which $State$ is optimal for the downstream LSTM cell. Formally, we construct a $State$ set $S_K$, which preserves the $K$ recently obtained $State$, and the maximum size $K$ is set ahead of time. The agent takes an action by sampling an optimal $State$ in $S_K$ from a multinomial distribution $\pi_K(k|s_t)$ as follows:
\begin{equation}
\begin{aligned}
P &= softmax(MLP(s_t)) \\
\pi_K(k|s_t) &= Pr(K=k|s_t) = \prod_{i=1}^Kp_i^{[k=i]},
\end{aligned}
\end{equation} 
where $[k=i]$ evaluates to 1 if $k=i, 0$ otherwise. $MLP$ represents a multilayer perceptron to transform $s_t$ to a vector with dimensionality $K$, and the $softmax$ function is used to transform the vector to a probability distribution $P$. $p_i$ is the $i$-th element in $P$. Then, the $State_{t-k}$ is transferred to the LSTM cell for further computation.

\noindent \textbf{Reward function.} Reward function is an indicator of the skip utility. A suitable reward function could guide the agent to select a series of optimal skip actions for training a better predictor. We capture this intuition by setting the reward to be the predicted log-likelihood of the ground truth, i.e., $R = \log Pr(\hat{y}_T|h_T)$. Therefore, by interacting with the environment through the rewards, the agent is incentivized to select the optimal skips to promote the probability of the ground truth.

\begin{figure}[t]
\centering
  \includegraphics[width=2.5in]{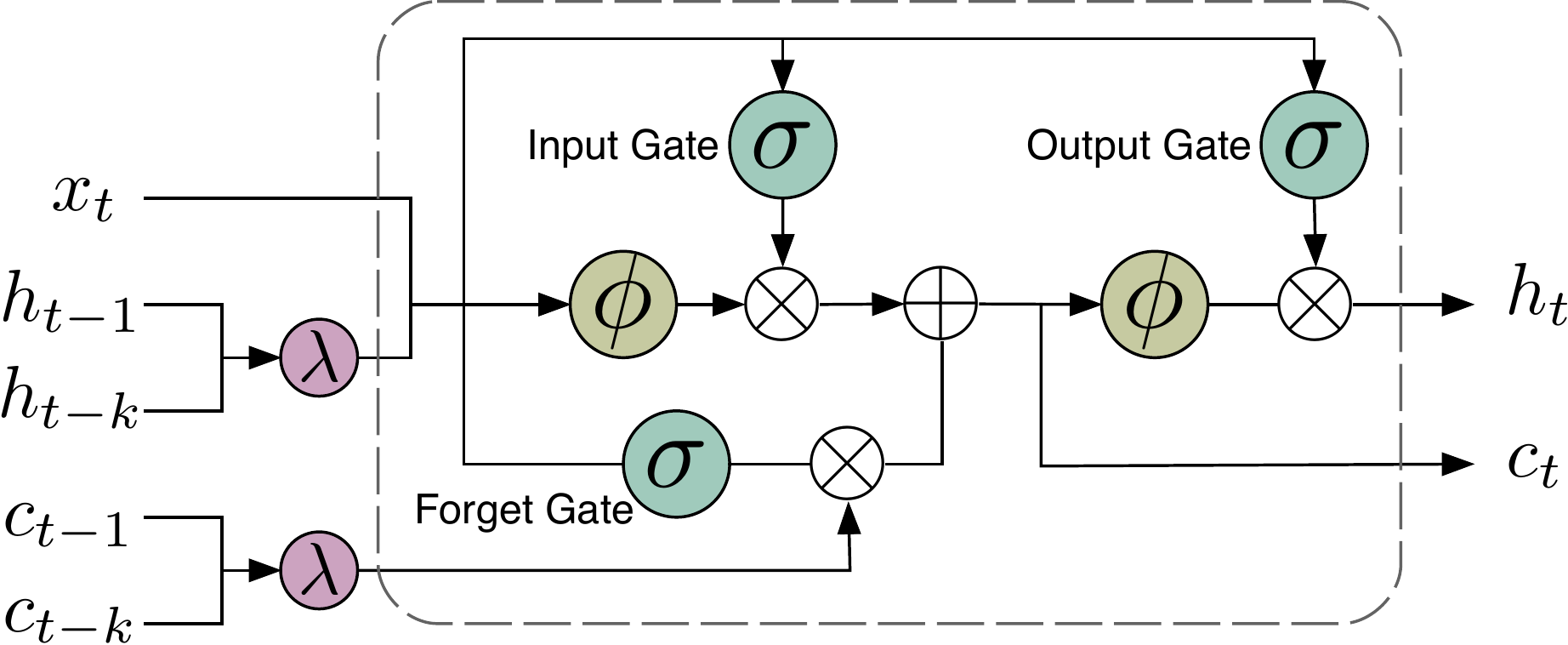}
  \caption{Schematic of the recurrent transition function encoding both the selected hidden/cell states and the previous hidden/cell states. $\lambda$ refers to the function $\lambda a + (1-\lambda)b$. $\sigma$ and $\phi$ refer to the sigmoid and tanh functions, respectively.} \label{fig:LSTM}
\end{figure}

\noindent \textbf{Recurrent transition function.} Based on the previously mentioned technique, we use a standard LSTM cell to encode the selected $State_{t-k}$, where $k\in\{1,2,...,K\}$. In the text classification experiments, we found that adding the additional immediate previous state $State_{t-1}$ usually led better results, although $State_{t-1}$ is a particular case of $State_{t-k}$. However, in our sequence labeling tasks, we found that just using $State_{t-k}$ is almost the optimal solution. Therefore, In our model, we use a hyperparameter $\lambda$ to incorporate these two situations, as shown in Figure~\ref{fig:LSTM}. Formally, we give the LSTM function as follows:
\begin{equation}
\begin{aligned}
  \mathbf{\widetilde{h}}_{t-1} &= \lambda\mathbf{h}_{t-k} + (1-\lambda)\mathbf{h}_{t-1} \\
  \mathbf{\widetilde{c}}_{t-1} &= \lambda\mathbf{c}_{t-k} + (1-\lambda)\mathbf{c}_{t-1} \\
  \begin{bmatrix}
  {\mathbf{g}}_t\\
  {\mathbf{i}}_t\\
  {\mathbf{f}}_t\\
  {\mathbf{o}}_t
  \end{bmatrix}
  &=
  \begin{pmatrix}
    \begin{bmatrix}
    \mathbf{W}_x^{g},\mathbf{W}_{h}^{g}\\
    \mathbf{W}_x^{i},\mathbf{W}_{h}^{i}\\
    \mathbf{W}_x^{f},\mathbf{W}_{h}^{f}\\
    \mathbf{W}_x^{o},\mathbf{W}_{h}^{o}
    \end{bmatrix}
    \bullet
    \begin{bmatrix}
    \mathbf{x}_t\\
    \mathbf{\widetilde{h}}_{t-1}
    \end{bmatrix}
  +
    \begin{bmatrix}
    \mathbf{b}^{g}\\
    \mathbf{b}^{i}\\
    \mathbf{b}^{f}\\
    \mathbf{b}^{o}
    \end{bmatrix}\label{eq:lstm1}
  \end{pmatrix}\\
  \mathbf{c}_t\ \ \ &= \phi(\mathbf{g}_t) \odot \sigma(\mathbf{i}_t) + \widetilde{\mathbf{c}}_{t-1}\odot \sigma(\mathbf{f}_t) \\
  \mathbf{h}_t\ \ \ &= \sigma(\mathbf{o}_t)  \odot \phi\left( \mathbf{c}_t \right),
\end{aligned}
\end{equation}
where $k\in\{1,2,...,K\}$. $\phi$ is the $tanh$ operator, and $\sigma$ is the $sigmoid$ operator. $\odot$ and $\bullet$ represent the Hadamard product and the matrix product, respectively. We assume that $y$ is one of $\{g,i,f,o\}$. The LSTM has $N_h$ hidden units and $N_x$ is the dimensionality of word representation $\mathbf{x}_i$. Then, $\mathbf{W}_x^y \in \mathbb{R}^{N_h\times{N_x}}$, $\mathbf{W}_h^y \in \mathbb{R}^{N_h\times{N_h}}$, and $\mathbf{b}^y \in \mathbb{R}^{N_h}$ are the parameters of the standard LSTM cell.


\begin{table*}[t]
\centering
\begin{tabular}{cccccccc}
\hline
  Task & Dataset & Level & Vocab  & \textbf{\#}Train & \textbf{\#}Dev & \textbf{\#}Test & \textbf{\#}class \\
  \hline
  Named Entity Recognition & CoNLL2003 & word & 30,290 & 204,567 & 51,578 & 46,666 & 17 \\
  Language Modeling & Penn Treebank & word & 10K & 929,590 & 73,761 & 82,431 & 10K \\
  Sentiment Analysis & IMDB & sentence & 112,540 & 21,250 & 3,750 & 25,000 & 2 \\
  Number Prediction & synthetic & word & 10 & 100,000 & 10,000 & 10,000 & 10 \\
  \hline 
\end{tabular}
\caption{Statistics of the CoNLL2003, Penn Treebank, IMDB, and synthetic datasets.}
  \label{tab:statistics}
\end{table*}

\subsection{Training Methods}
Our goal for training is optimizing the parameters of the policy gradient agent $\theta_a$, together with the parameters of standard LSTM and possibly other parameters denoted as $\theta_l$. Optimizing $\theta_l$ is straightforward and can be treated as a classification problem. Because the cross entropy loss $J_1(\theta_l)$ is differentiable, we can apply backpropagation to minimize it as follows:
\begin{equation}
J_1(\theta_l) = -[y_i\log{\hat{y}_i}+(1-y_i)\log(1-\hat{y}_i)],
\end{equation}
where $\hat{y}_i$ is the output of the model.

The objective of training the agent is to maximize the expected reward under the skip policy distribution plus an entropy regularization~\cite{nachum2017bridging}.
\begin{equation}
J_2(\theta_a) = \mathbb{E}_{\pi(a_{1:T})}[R] + H(\pi(a_{1:T})),
\end{equation}
where $\pi(a_{1:T}) = \prod_{t=1}^T Pr(a_t|s_t;\theta_a)$, and $R = \log Pr(\hat{y}_T|h_T)$. $H(\pi(a_{1:T})) =-\mathbb{E}_{\pi(a_{1:T})}[\log \pi(a_{1:T})]$ is an entropy term, which can prevent premature entropy collapse and encourage the policy to explore more diverse space. We provide evidence that using the reinforcement learning with an entropy term can model sentence better than attention-based connections, as shown in Appendix.

Because of the non-differentiable nature of discrete skips, we adopt a policy gradient formulation referred to as REINFORCE method~\cite{williams1992simple} to optimize $\theta_a$:
\begin{equation}
\begin{aligned}
\nabla_{\theta_a} J_2(\theta_a) = &\mathbb{E}_{\pi(a_{1:T})}[\sum_{t=1}^T\nabla_{\theta_a} \log Pr(a_t|s_t;\theta_a)* \\
&(R - \sum_{t=1}^T\log Pr(a_t|s_t;\theta_a) - 1)].\\
\end{aligned}
\end{equation}
By applying the above algorithm, the loss $J_2(\theta_a)$ can be computed by standard backpropagation. Then, we can get the final objective by minimizing the following function:
\begin{equation}
J(\theta_a,\theta_l) = \frac{1}{M}[\sum_{m=1}^M(J_1(\theta_l)-J_2(\theta_a))],
\end{equation}
where $M$ denotes the quantity of the minibatch, and the objective function is fully differentiable.

\section{Experiments and Results}
In this section, we present the experimental results of the proposed model for a variety of sequence modeling tasks, such as named entity recognition, language modeling, and sentiment analysis. In addition to the evaluation matrices for each task, in order to better understand the advantages of our model, we visualize the behavior of skip actions and make a comparison about how the gradients get changed between the LSTM and the proposed model. In addition, we also evaluate the model on the synthetic number prediction tasks, and verify the proficiency in long-term dependencies. The datasets used in the experiments are listed in Table~\ref{tab:statistics}.

\noindent \textbf{General experiment settings.} For the fair comparison, we use the same hyperparameters and optimizer with each baseline model of different tasks, which will be detailed in each experiment. As for the policy gradient agent, we use single layer MLPs with 50 hidden units. The maximum size of skip $K$ and the hyperparameter $\lambda$ are fixed during both training and testing. 

\begin{table}[t]
\centering
\scalebox{0.95}{
\begin{tabular}{l|c}
\hline
  \textbf{Model}  & \textbf{F1} \\
  \hline
  \citeauthor{huang2015bidirectional} \shortcite{huang2015bidirectional}& 90.10\\
  \citeauthor{chiu2015named} \shortcite{chiu2015named}& 90.91$\pm$0.20\\
  \citeauthor{lample2016neural} \shortcite{lample2016neural}& 90.94\\
  \citeauthor{ma2016end} \shortcite{ma2016end}& 91.21 \\
  \citeauthor{strubell2017fast} \shortcite{strubell2017fast}$\dagger$ & 90.54 $\pm$ 0.18 \\
  \citeauthor{strubell2017fast} \shortcite{strubell2017fast} & 90.85 $\pm$ 0.29 \\
  \hline
  LSTM, fixed skip = 3~\cite{zhang2016architectural} & 91.14 \\
  LSTM, fixed skip = 5~\cite{zhang2016architectural} & 91.16 \\
  LSTM with attention & 91.23 \\
  LSTM with dynamic skip & \textbf{91.56} \\
  \hline
\end{tabular}}
\caption{F1-measure of different methods applied to the CoNLL 2003 dataset. The model that does not use character embeddings is marked with $\dagger$. ``LSTM with attention'' refers to the LSTM model using attention mechanism to connect two words.}
  \label{tab:ner}
\end{table}

\begin{table*}[t]
\centering
\begin{tabular}{l|c|c|c}
\hline
  \textbf{Model}  & \textbf{Dev.}(\textit{PPL}) & \textbf{Test}(\textit{PPL}) & \textbf{Size}\\
  \hline
  RNN~\cite{mikolov2012context} & - & 124.7 & 6 m\\
  RNN-LDA~\cite{mikolov2012context} & - & 113.7 & 7 m\\
  Deep RNN~\cite{pascanu2013construct} & - & 107.5 & 6 m\\
  Zoneout + Variational LSTM (medium)~\cite{merity2016pointer}$\dagger$ & 84.4 & 80.6 & 20 m\\
  Variational LSTM (medium)~\cite{gal2016theoretically}$\dagger$ & 81.9 & 79.7 & 20 m\\
  Variational LSTM (medium, MC)~\cite{gal2016theoretically}$\dagger$ & - & 78.6 & 20 m\\
  \hline
  \hline
  Regularized LSTM~\cite{zaremba2014recurrent}$\dagger$$\ddagger$ & 86.2 & 82.7 & 20 m \\
  Regularized LSTM, fixed skip = 3~\cite{zhang2016architectural}$\dagger$ & 85.3 & 81.5 & 20 m \\
  Regularized LSTM, fixed skip = 5~\cite{zhang2016architectural}$\dagger$ & 86.2 & 82.0 & 20 m \\
  Regularized LSTM with attention$\dagger$ & 85.1 & 81.4 & 20 m \\
  Regularized LSTM with dynamic skip, $\lambda$=1, K=5$\dagger$ & \textbf{82.5} & \textbf{78.5} & 20 m \\
  \hline
  \hline
  CharLM~\cite{kim2016character}$\dagger$$\ddagger$ & 82.0 & 78.9 & 19 m \\
  CharLM, fixed skip = 3~\cite{zhang2016architectural}$\dagger$ & 83.6 & 80.2 & 19 m \\
  CharLM, fixed skip = 5~\cite{zhang2016architectural}$\dagger$ & 84.9 & 80.9& 19 m \\
  CharLM with attention$\dagger$ & 82.2 & 79.0 & 19 m \\
  CharLM with dynamic skip, $\lambda$=1, K=5$\dagger$ & \textbf{79.9} & \textbf{76.5} & 19 m \\
  \hline
\end{tabular}
\caption{Perplexity on validation and test sets for the Penn Treebank language modeling task. \textit{PPL} refers to the average perplexity (lower is better) in ten runs. Size refers to the approximate number of parameters in the model. The models marked with $\dagger$ have the same configuration which features a hidden size of 650 and a two layer LSTM. The models marked with $\ddagger$ are equivalent to the proposed model with hyperparameters $\lambda$ = 0, and $K$ = 1.}
  \label{tab:LM}
\end{table*}

\subsection{Named Entity Recognition}
We now present the results for a sequence modeling task, Named Entity Recognition (NER). We performed experiments on the English data from CoNLL 2003 shared task~\cite{tjong2003introduction}. This data set contains four different types of named entities: locations, persons, organizations, and miscellaneous entities that do not belong in any of the three previous categories. The corpora statistics are shown in Table~\ref{tab:statistics}. We used the BIOES tagging scheme instead of standard BIO2, as previous studies have reported meaningful improvement with this scheme~\cite{lample2016neural,ma2016end}.


Following~\cite{ma2016end}, we use 100-dimension GloVe word embeddings\footnote{ http://nlp.stanford.edu/projects/glove/} and unpretrained character embeddings as initialization. We use a forward and backward new LSTM layer with $\lambda=1, K=5$ and a CRF layer to achieve this task. The reward is probability of true label sequence in CRF. We use early stopping based on performance on validation sets. \citeauthor{ma2016end} \shortcite{ma2016end} reported the “best” result appeared at 50 epochs and the model training required 8 hours. In our experiment, because of more exploration, the “best” result appeared at 65 epochs, the proposed method training required 9.98 hours.

\begin{figure}[t]
\centering
  \includegraphics[width=2.5in]{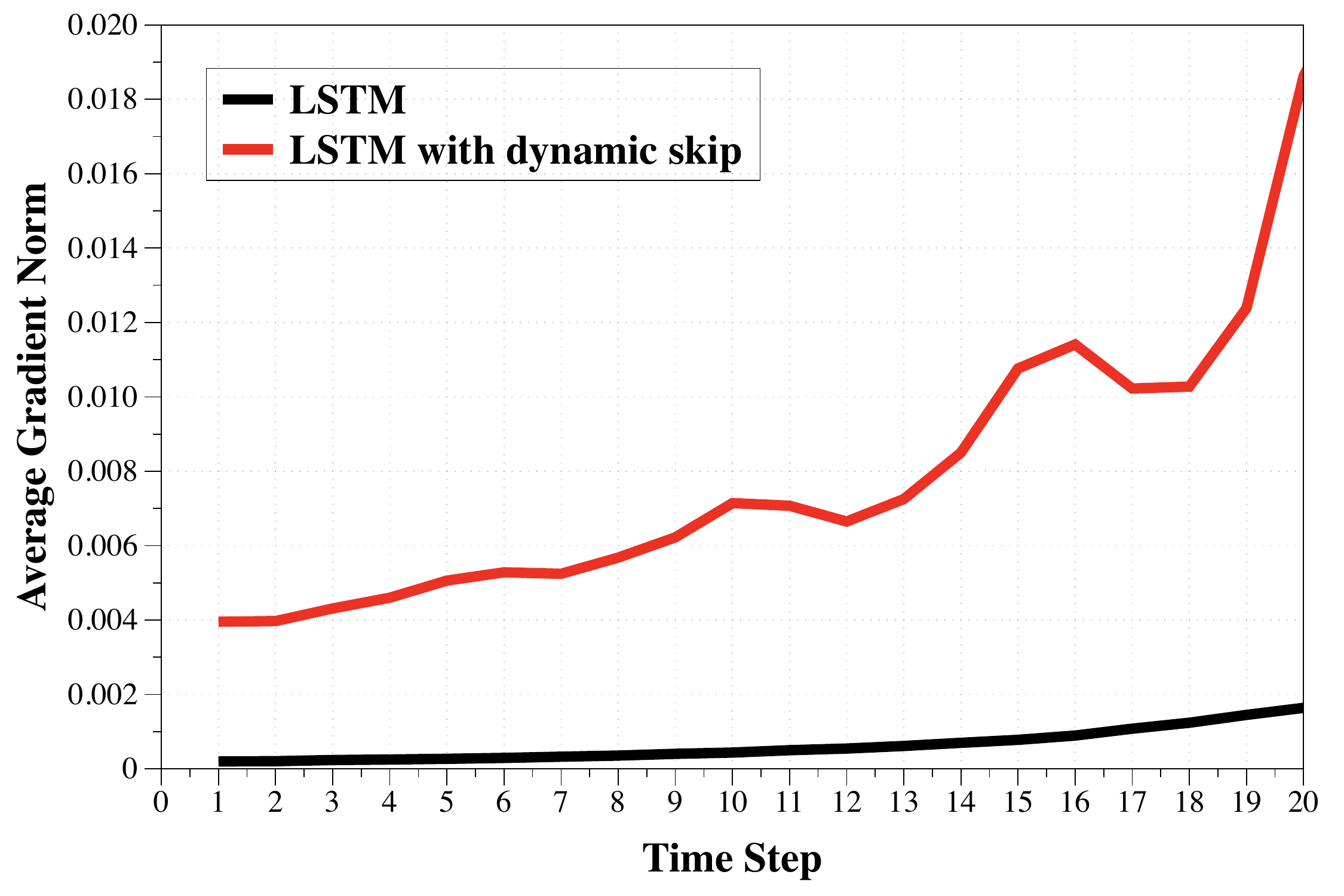}
  \caption{Normalized long-term gradient values $\left \| \frac{\partial L_T}{\partial h_t} \right \|$ tested on CoNLL 2003 dataset. At the initial time steps, the proposed model still preserves effective gradients, which is hundreds of times larger than those in the standard LSTM, indicating that the proposed model have stronger ability to capture long-term dependency.} \label{fig:gradient}
\end{figure}

Table~\ref{tab:ner} shows the F1 scores of previous models and our model for NER on the test dataset from the CoNLL 2003 shared task. To our knowledge, the previous best F1 score (91.21) was achieved by using a combination of bidirectional LSTM, CNN, and CRF to obtain both word- and character-level representations automatically~\cite{ma2016end}. By adding a bidirectional LSTM with a fixed skip, the model's performance would not improve. By contrast, the model using dynamic skipping technique improves the performance by average of 0.35\%, and error reduction rate is more than 4\%. Our model also outperforms the attention model, because the attention model employs a deterministic network to compute an expectation over the alignment variable, i.e., $\log f(H,\mathbb{E}_k[k])$, not the expectation over the features, i.e., $\log\mathbb{E}_k(H,k)$. However, the gap between the above two expectations may be large~\cite{deng2018latent}. Our model is equivalent to the ``Categorical Alignments'' model in \cite{deng2018latent}, which can effectively tackle the above deficiency. The detailed proof is given in Appendix.


In Figure~\ref{fig:gradient}, we investigate the problem of vanishing gradient by comparing the long-term gradient values between standard LSTM and the proposed model. Following~\cite{mujika2017fast}, we compute the average gradient norms $\left \| \frac{\partial L_T}{\partial h_t} \right \|$ of the loss at time $T$ with respect to hidden state $h_t$ at each time step $t$. We visualize the gradient norms in the first 20 time steps by normalizing the average gradient norms by the sum of average gradient norms for all time steps. Evidently, at the initial time steps, the proposed model still preserves effective gradient backpropagation. The average gradient norm in the proposed model is hundreds of times larger than that in the standard LSTM, indicating that the proposed model captures long-term dependencies, whereas the standard LSTM basically stores short-term information~\cite{mujika2017fast}. The same effect was observed for cell states $c_t$.

\subsection{Language Modeling}
We also evaluate the proposed model on the Penn Treebank language model corpus~\cite{marcus1993building}. The corpora statistics are shown in Table~\ref{tab:statistics}. The model is trained to predict the next word (evaluated on perplexity) in a sequence.

To exclude the potential impact of advanced models, we restrict our comparison among the RNNs models. We replicate settings from Regularized LSTM~\cite{zaremba2014recurrent} and CharLM~\cite{kim2016character}. The above networks both have two layers of LSTM with 650 units, and the weights are initialized uniformly [-0.05, +0.05]. The gradients backpropagate for 35 time steps using stochastic gradient descent, with a learning rate initially set to 1.0. The norm of the gradients is constrained to be below five. Dropout with a probability of 0.5 on the LSTM input-to-hidden layers and the hidden-to-output softmax layer is applied. The main difference between the two models is that the former model uses word embeddings as inputs, while the latter relies only on character-level inputs.

We replace the second layer of LSTM in the above baseline models with a fixed skip LSTM or our proposed model. The testing results are listed in Table~\ref{tab:LM}. We can see that the performance of the LSTM with a fixed skip may be even worse than that of the standard LSTM in some cases. This verifies that in some simple tasks such as the adding problem, copying memory problem, and sequential MNIST problem, the LSTM with a fixed skip length may be quite powerful~\cite{zhang2016architectural}, whereas in a complex language environment, the fixed skip length is constrained by its inability to take advantage of the dependencies with variable lengths.

\begin{figure}[t]
\centering
  \includegraphics[width=3.4in]{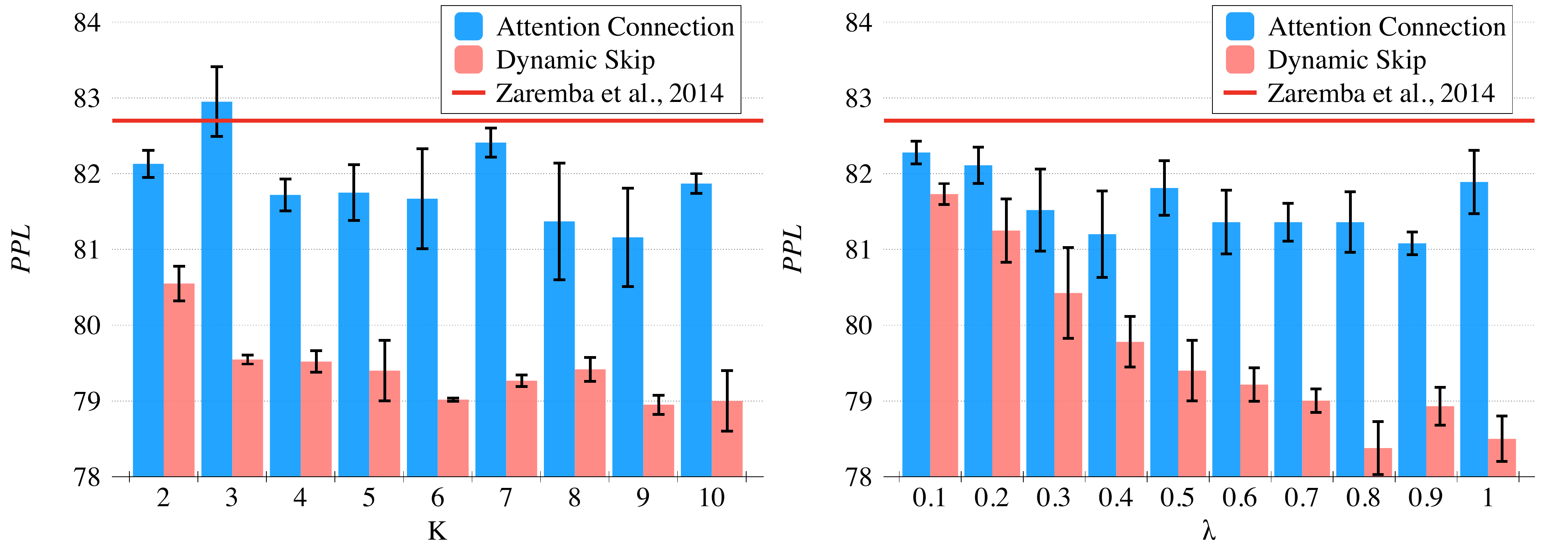}
  \caption{Test set perplexities (lower is better) on Penn Tree-bank language model corpus with standard deviation for $K$ from 2 to 10 with $\lambda=0.5$ (left), and $\lambda$ from 0.1 to 1.0 with $K = 5$ (right).} \label{fig:LMlambda}
\end{figure}

Hence, by adding a dynamic skip on the recurrent connections to the LSTM, our model can effectively tackle the above problem. Both the models with dynamic skip connections outperform baseline models, and the best model is able to improve the average test perplexity from 82.7 to 78.5, and 78.9 to 76.5, respectively. We also investigated how the hyperparameters $\lambda$ and $K$ affect the performance of proposed model as shown in Figure~\ref{fig:LMlambda}. $\lambda$ is a weight to balance the utility between the newly selected $State$ and the previous $State$. $K$ represents the size of the skip space. From both figures we can see that, in general, the proposed model with larger $\lambda$ and $K$ values should obtain better \textit{PPL}, while producing a larger variance because of the balance favoring the selected $State$ and a larger searching space.

\begin{figure}
\centering
  \includegraphics[width=2.8in]{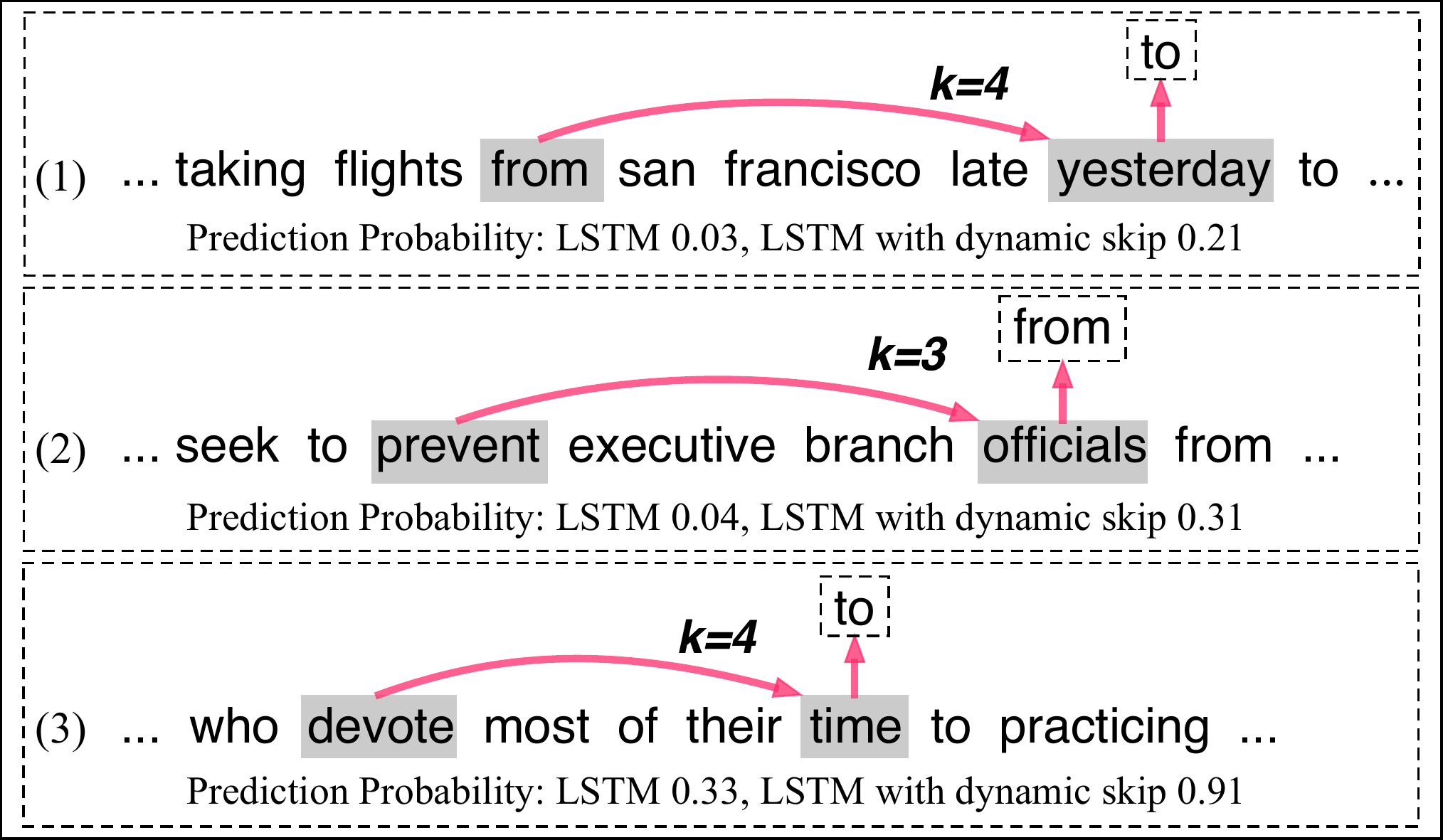}
  \caption{Examples of the proposed model applied to language modeling.} \label{fig:LM}
\end{figure}

To verify the effectiveness of dynamic skip connections, we visualize the behavior of the agent in a situation where it predicts a preposition based on a long-term dependency. Some typical examples are shown in Figure~\ref{fig:LM}. From the figure, we can see that in the standard LSTM setting, predicting the three prepositions in the figure is difficult just based on the previous $State$. By introducing the long-term and the most relevant $State$ using dynamic skip connections, the proposed model is able to predict the next word with better performance in some cases.

\begin{table}[t]
\centering
\scalebox{0.95}{
\begin{tabular}{l|c}
\hline
  \textbf{Model} & \textbf{Acc.} \\
  \hline
  LSTM & 89.1 \\
  LSTM + LA~\cite{chen2016neural} & 89.3\\
  LSTM + CBA$^G$~\cite{long2017cognition} & 89.4\\
  LSTM + CBA + LA$^G_s$~\cite{long2017cognition} & 89.8\\
  LSTM + CBA + LA$^G_p$~\cite{long2017cognition} & \textbf{90.1}\\
  Skip LSTM~\cite{campos2017skip} & 86.6\\
  Jump LSTM~\cite{yu2017learning} & 89.4\\
  \hline
  LSTM, fixed skip = 3~\cite{zhang2016architectural} & 89.6\\
  LSTM, fixed skip = 5~\cite{zhang2016architectural} & 89.3\\
  LSTM with attention & 89.4\\
  LSTM with dynamic skip, $\lambda$=0.5, K=3 & \textbf{90.1}\\
  \hline
\end{tabular}}
\caption{Accuracy on the IMDB test set.}
  \label{tab:sentiment}
\end{table}

\subsection{Sentiment Analysis on IMDB}
Two similar works to ours are Jump LSTM~\cite{yu2017learning} and Skip LSTM~\cite{campos2017skip}, which are excellent modifications to the LSTM and achieve great performance on text classification. However, the above model cannot be applied to some sequence modeling tasks, such as language modeling and named entity recognition, because the jumping characteristics and the nature of skipping state updates make the models cannot produce LSTM outputs for skipped tokens and cannot update the hidden state, respectively. For better comparison with the above two models, we apply the proposed model to a sentiment analysis task.

The IMDB dataset~\cite{maas2011learning} contains 25,000 training and 25,000 testing movie reviews annotated into \textit{positive} or \textit{negative} sentiments, where the average length of text is 240 words. We randomly set aside about 15\% of the training data for validation. The proposed model, Jump LSTM, Skip LSTM, and LSTM all have one layer and 128 hidden units, and the batch size is 50. We use pretrained word2vec embeddings as initialization when available, or random vectors drawn from $\mathcal{U}(-0.25,+0.25)$. Dropout with a rate of 0.2 is applied between the last LSTM state and the classification layer. We either pad a short sequence or crop a long sequence to 400 words. We set $\lambda$ and $K$ to 0.5 and 3, respectively.

The result is reported in Table~\ref{tab:sentiment}. \cite{chen2016neural} employed the idea of local semantic attention to achieve the task. \cite{long2017cognition} proposed a cognition based attention model, which needed additional eye-tracking data. From this result, we can see that our model also exhibits a strong performance on the text classification task. The proposed model is 1\% better than the standard LSTM model. In addition, our model outperforms Skip LSTM and Jump LSTM models with accuracy at 90.1\%. Therefore, the proposed model not only can achieve sequence modeling tasks such as language modeling and named entity recognition, but it also has a stronger ability for text classification tasks than Jump LSTM and Skip LSTM.

\begin{table}[t]
\centering
\begin{tabular}{|l|c|c|}
\hline
  \multicolumn{3}{|c|}{sequence length 11}\\
  \hline
  \textbf{Model}  & \textbf{Dev.} & \textbf{Test}\\
  \hline
  LSTM & 69.6 & 70.4\\
  LSTM with attention & 71.3 & 72.5\\
  LSTM with dynamic skip, $\lambda$=1, \ \ \ K=10 & 79.6 & 80.5\\
  LSTM with dynamic skip, $\lambda$=0.5, K=10 & \textbf{90.4} & \textbf{90.5}\\
  \hline
  \multicolumn{3}{|c|}{sequence length 21}\\
  \hline
  \textbf{Model}  & \textbf{Dev.} & \textbf{Test}\\
  \hline
  LSTM & 26.2 & 26.4\\
  LSTM with attention & 26.7 & 26.9\\
  LSTM with dynamic skip, $\lambda$=1, \ \ \ K=10 & 77.6 & 77.7\\
  LSTM with dynamic skip, $\lambda$=0.5, K=10 & \textbf{87.7} & \textbf{88.5}\\
  \hline
\end{tabular}
\caption{Accuracies of different methods on number prediction dataset.}
  \label{tab:number}
\end{table}

\subsection{Number Prediction with Skips}
For further verification that the Dynamic LSTM is indeed able to learn how to skip if a clear skipping signal is given in the text, similar to~\cite{yu2017learning}, we also investigate a new task, where the network is given a sequence of $L$ positive integers $x_{0:T-1}$, and the label is $y=x_{x_{T-1}}$. Here are two examples to illustrate the idea:
\begin{table}[h]
\centering
\begin{tabular}{c}
  input1: 8, 5, 1, \textbf{\underline7}, 4, \underline3. \ label: 7\\
  input2: 2, 6, \textbf{\underline4}, 1, 3, \underline2. \ label: 4
\end{tabular}
\end{table}

As the examples show, $x_{T-1}$ is the skipping signal that guides the network to introduce the $x_{T-1}$-th integer as the input to predict the label. The ideal network should learn to ignore the remaining useless numbers and learn how to skip from the training data.

According to above rule, we generate 100,000 training, 10,000 validation, and 10,000 test examples. Each example with a length $T=11$ is formed by randomly sampling 11 numbers from the integer set $\{0,1,...,9\}$, and we set $x_{x_{11}}$ as the label of each example. We use ten dimensional one-hot vectors to represent the integers as the sequential inputs of LSTM or Dynamic LSTM, of which the last hidden state is used for prediction. We adopt one layer of LSTM with 200 hidden neurons. The Adam optimizer~\cite{kingma2014adam} trained with cross-entropy loss is used with 0.001 as the default learning rate. The testing result is reported in Table~\ref{tab:number}. It is interesting to see that even for a simple task, the LSTM model cannot achieve a high accuracy. However, the LSTM with dynamic skip is able to learn how to skip from the training examples to achieve a much better performance.

Taking this one step further, we increase the difficulty of the task by using two skips to find the label, i.e., the label is $y=x_{x'} , x'=x_{x_{T-1}}$. To accord with the nature of skip, we force $x'<x_{T-1}$, Here is an example:
\begin{table}[h]
\centering
\begin{tabular}{c}
  input: 8, \textbf{\underline5}, 1, 7, \underline1, 3, 3, 4, 7, 9, \underline4. \ label: 5
\end{tabular}
\end{table}

Similar to the former method, we construct a dataset with the same size, 100,000 training, 10,000 validation, and 10,000 test examples. Each example with length $T=21$ is also formed by randomly sampling 21 numbers from the integer set $\{0, 1, ..., 9\}$. We use the same model trained on the dataset. As the Table~\ref{tab:number} shows, the accuracy of the LSTM with dynamic skip is vastly superior to that of LSTM. Therefore, the results indicate that the Dynamic LSTM is able to learn how to skip.

\section{Related Work}
Many attempts have been made to overcome the difficulties of RNNs in modeling long sequential data, such as gating mechanism~\cite{hochreiter1997long,chung2014empirical}, Multi-timescale mechanism~\cite{chung2016hierarchical}. Recently, many works have explored the use of skip connections across multiple time steps~\cite{zhang2016architectural,chang2017dilated}. \citeauthor{zhang2016architectural} \shortcite{zhang2016architectural} introduced the \textit{recurrent skip coefficient}, which captures how quickly the information propagates over time, and found that raising the recurrent skip coefficient can mostly improve the performance of models on long-term dependency tasks. Note that previous research on skip connections all focused on a fixed skip length, which is set in advance. Different from these methods, this work proposed a reinforcement learning method to dynamically decide the skip length.


Other relevant works that introduce reinforcement learning to recurrent neural networks are Jump LSTM~\cite{yu2017learning}, Skip RNN~\cite{seo2017neural}, and Skim RNN~\cite{seo2017neural}. The Jump LSTM aims to reduce the computational cost of RNNs by skipping irrelevant information if needed. Their model learns how many words should be omitted, which also utilizes the REINFORCE algorithm. Also, the Skip RNN and Skim RNN can learn to skip (part of) state updates with a fully differentiable method. The main differences between our method and the above methods are that Jump LSTM can not produce LSTM outputs for the skipped tokens and the Skip (Skim) RNN would not update (part of) hidden states. Thus three models would be difficult to be used for sequence labeling tasks. In contrast to them, our model updated the entire hidden state at each time step, and can be suitable for sequence labeling tasks.


\section{Conclusions}
In this work, we propose a reinforcement learning-based LSTM model that extends the existing LSTM model with dynamic skip connections. The proposed model can dynamically choose one optimal set of hidden and cell states from the past few states. By means of the dynamic skip connections, the model has a stronger ability to model sentences than those with fixed skip, and can tackle the dependency problem with variable lengths in the language. In addition, because of the shorter gradient backpropagation path, the model can alleviate the challenges of vanishing gradient. Experimental results on a series of sequence modeling tasks demonstrate that the proposed method can achieve much better performance than previous methods.

\section{Acknowledgments}
The authors wish to thank the anonymous reviewers for their helpful comments. This work was partially funded by China National Key R\&D Program (No. 2017YFB1002104, 2018YFC0831105), National Natural Science Foundation of China (No. 61532011,61751201, 61473092, and 61472088), and STCSM (No.16JC1420401, 17JC1420200).

\section{Appendix}

\subsection{Notations}
Let $H \in \mathbb{R}^{d\times T}$ be an matrix formed by a set of members $\{h_1, h_2, \dots, h_T \}$, where $h_t \in \mathbb{R}^d$ is vector-valued and $T$ is the cardinality of the set. Let $s_t$ be an arbitrary ``query'' for attention computation or the state of reinforcement learning agent. In the paper, the $s_t$ is defined by incorporating the previous hidden state $h_{t-1}$ and the current input ${x_t}$ as follows:
$$s_t = h_{t-1} \oplus x_t.$$

Then, we use $s_t$ to operate on $H$ to predict the label $y \in \mathcal{Y}$. The process can be formally defined as follows:
\begin{equation}
\begin{aligned}
z &= \mathcal{D}[g(H, s_t; \theta)] \\
y &= f(H, z; \theta), \\
\end{aligned}
\end{equation}
where $g$ is a function to produce an alignment distribution $\mathcal{D}$. $f$ is another function mapping $H$ over the distribution $z$ to the label $y$. Our goal is to optimize the parameters $\theta$ by maximizing the log marginal likelihood:
\begin{equation}
\begin{aligned}
\max\limits_{\theta}\log p(y&=\hat{y}|H, s_t) \\
&= \max\limits_{\theta} \log \mathbb{E}_{z}[f(H,z;\theta)] \\
&= \max\limits_{\theta} \log \int_{z} q(z|s_t;\theta)f(H,z;\theta)d_z.
\end{aligned}
\end{equation}
Directly maximizing this log marginal likelihood is often difficult due to the expectation~\cite{deng2018latent}. To tackle this challenge, previous works focus on using the attention model as an alternative solution.

\subsection{Deficiency of Attention model}
Attention networks use a deterministic network to compute an expectation over the distribution variable $z$. We can write this model as follows:
\begin{equation}
\begin{aligned}
\log p_{att}(y=\hat{y}|H,s_t)&=\log p_{att}(y=\hat{y}|\mathbb{E}_z[H],s_t)\\
&=\log f(H, \mathbb{E}_z[z]). \\
\end{aligned}
\end{equation}
The attention networks compute the expectation before $f$ without computing an integral over $f$, which enhance the efficiency of computation. Although many works use attention as an approximation of alignment~\cite{cohn2016incorporating,tu2016modeling}, some works also find that the attention model is not satisfying in some cases\cite{xu2015show}, because of depending on $f$, the gap between $\mathbb{E}_z[f(H,z)]$ and $f(H,\mathbb{E}_z[z])$ may be large~\cite{deng2018latent}.

\subsection{REINFORCE with Entropy Regularization}
In the previous section, we have show that the log-probability of label $y$ can be mediated through a latent alignment variable $z$:
\begin{equation}
\log p(y=\hat{y}|H, s_t) = \log \mathbb{E}_{z}[f(H,z;\theta)].
\end{equation}
Through variational inference, the above function can be rewritten as:

\begin{equation}
\begin{aligned}
\log p(y|H,s_t) =& \int q_\phi(z|H,s_t)\log \frac{p_\theta(z,y)}{p_\theta(z|y)} dz\\
=& \int q_\phi(z|H,s_t)\log\frac{p_\theta(z,y)}{q_\phi(z|H,s_t)} \\
&+ KL[q_\phi(z|H,s_t)||p_\theta(z|y)],
\end{aligned}
\end{equation}
The second term is the KL divergence of the approximate from the true posterior. Since this KL-divergence is non-negative, the first term is called the (variational) \textit{lower bound} $\mathcal{L}(\theta, \phi; y)$ and can be written as:
\begin{equation}
\begin{aligned}
\log p(y|H,s_t) \geq &\mathcal{L}(\theta,\phi|H,s_t)\\
\mathcal{L}(\theta,\phi|H,s_t) = & \int q_\phi(z|H,s_t)\log p_\theta(y|z)dz\\
& +\int q_\phi(z|H,s_t)\log \frac{p_\theta(z)}{q_\phi(z|H,s_t)}dz, 
\end{aligned}
\end{equation}
which can also be written as:
\begin{equation}
\begin{aligned}
\log p(y|H,s_t) \geq  & \ \  \mathbb{E}_{q_\phi(z|H,s_t)}[\log p_\theta(y|z)]\\
&-KL[q_\phi(z|H,s_t)||p_\theta(z)],
 \end{aligned}
\end{equation}
where the first term is the prediction loss, or expected log-likelihood of the label. The expectation is taken with respect to the encoder’s distribution over the representations. This term encourages the decoder to learn to predict the true label. The second term is a regularizer. This is the KL divergence between the encoder’s distribution $q_\phi(z|H,s_t)$ and $p_\theta(z)$.

In order to tighten the gap between the lower bound and the likelihood of the label, we should maximize the variational lower bound:
\begin{equation}
\max_{\phi,\theta} \mathbb{E}_{q_\phi(z|H,s_t)}[\log p_\theta(y|z)]-KL[q_\phi(z|H,s_t)||p_\theta(z)].
\label{maximize}
\end{equation}

In our LSTM with dynamic skip setting, let the random variable $z$ be the trajectory variable $\tau$. Then the function~\ref{maximize} can be rewritten as follows:
\begin{equation}
\max_{\theta_a,\theta_l} \mathbb{E}_{q_{\theta_a}(\tau|H,s_t)}[\log p_{\theta_l}(y|\tau)]-KL[q_{\theta_a}(\tau|H,s_t)||p_{\theta_l}(\tau)],
\label{maxRL}
\end{equation}
where $\theta_a$ is the parameters of agent, and $\theta_l$ is the parameters of LSTM units. For simplicity, we omit the condition of $q_{\theta_a}(\tau|H,s_t)$ to be $q_{\theta_a}(\tau)$, and treat the log-likelihood of ground truth $\log p_{\theta_l}(y|\tau)$ as the rewards $R_{\theta_l}(\tau)$. We make the prior $p_\theta(z)$ be uniform distribution. Then the gradients of function~\ref{maxRL} can be computed as follows:
\begin{equation}
\begin{aligned}
\nabla J(\theta_a) = & \nabla \mathbb{E}_{q_{\theta_a}(\tau)}[R_{\theta_l}(\tau)]-KL[q_{\theta_a}(\tau)||p_{\theta_l}(\tau)] \\
= & \nabla \mathbb{E}_{q_{\theta_a}(\tau)}[R_{\theta_l}(\tau) - \log q_{\theta_a}(\tau)] \\
= & \nabla \int_\tau (R_{\theta_l}(\tau) - \log q_{\theta_a}(\tau))q_{\theta_a}(\tau)d\tau \\
= & \int_\tau (R_{\theta_l}(\tau) - \log q_{\theta_a}(\tau) -1)\nabla q_{\theta_a}(\tau)d\tau \\
=  \int_\tau q_{\theta_a}&(\tau)\nabla\log q_{\theta_a}(\tau)(R_{\theta_l}(\tau) - \log q_{\theta_a}(\tau) -1)d\tau \\
=& \mathbb{E}_{q_{\theta_a}(\tau)}[\sum_{t=1}^T\nabla_{\theta_a}\log\pi_{\theta_a}(a_t|s_t)]* \\
& (R_{\theta_l}(\tau)-\sum_{t=1}^T\log\pi_{\theta_a}(a_t|s_t)-1),
\end{aligned}
\end{equation}
which has the same form as the loss function in the paper. Hence, using REINFORCE with entropy regularization can optimize the model towards the direction of tightening the gap between the lower bound and the likelihood of the label, while the attention model may not tackle this gap. This is why the performance of REINFORCE with an entropy term is better than attention. Meanwhile, optimizing the parameters of standard LSTM $\theta_l$ is straightforward and can be treated as a classification problem as shown in our paper. Therefore, our model can be trained end-to-end with standard back-propagation.

\bibliography{aaai}
\bibliographystyle{aaai}

\end{document}